\documentclass[english]{ccdconf}
\begin{document}

\title{A Hierarchical Emotion Regulated Sensorimotor Model: Case Studies}
\author{Junpei Zhong\aref{ply,waseda}, Rony Novianto\aref{syn},  Mingjun Dai\aref{sz}, Xinzheng Zhang\aref{jinan}, Angelo Cangelosi\aref{ply}}

\affiliation[ply]{Centre for Robotics and Neural Systems,
Plymouth University, Plymouth, PL4 8AA, United Kingdom.
        \email{\{junpei.zhong, a.cangelosi\}@plymouth.ac.uk}}
\affiliation[waseda]{Laboratory for Intelligent Dynamics and Representation, Waseda University, Tokyo, Japan.       }
\affiliation[syn]{Centre of Quantum Computation and Intelligent Systems, University of Technology, Sydney, Australia.
        \email{rony@ronynovianto.com}}
\affiliation[sz]{Shenzhen Key Lab of Advanced Communication and Information Processing and Shenzhen Key Laboratory of Media Security, Shenzhen University, Shenzhen, China.
        \email{mjdai@szu.edu.cn}}
    \affiliation[jinan]{School of Electrical and Information Engineering, Jinan University, Zhuhai, China. \email{ee.xz.zhang@gmail.com}}

\maketitle

\begin{abstract}
Inspired by the hierarchical cognitive architecture and the perception-action model (PAM)~\cite{preston2002empathy}, we propose that the internal status acts as a kind of common-coding representation which affects, mediates and even regulates the sensorimotor behaviours. 
These regulation can be depicted in the Bayesian framework, that is why cognitive agents are able to  generate behaviours with subtle differences according to their emotion or recognize the emotion by perception.   
A novel recurrent neural network called recurrent neural network with parametric bias units (RNNPB) runs in three modes, constructing a two-level emotion regulated learning model, was further applied to testify this theory in two different cases.
\end{abstract}


\footnotetext{JZ and AC were supported by the EU project POETICON++ under grant agreement 288382 and UK EPSRC project BABEL. MD was supported by NSF of China (61301182),  NSF of Guangdong (S2013040016857), Specialized Research Fund for the Doctoral Program of Higher Education from the Ministry of Education (20134408120004), Yumiao Engineering from Education Department of Guangdong (2013LYM\_0077), Foundation of Shenzhen City (KQCX20140509172609163), and from NSF of Shenzhen University (00002501, 00036107). JZ would like to acknowledge the data-set and inspirations from L. Ca\~{n}amero and M. Lewis from University of Hertfordshire.}

\section{Introduction}
It is widely agreed that the internal state interacts with, modulates and mediates various cognitive processes including sensorimotor process, 
in the sense that the sensorimotor expression, for instance, the physical expression as motor primitives, are partially caused by, or co-occur with, emotions. 
For instance,~\cite{argyle2013bodily,darwin1998expression,hadjikhani2003seeing,huis2014body} found that body languages have been found to constitute a signal of affective information. 
certain subtle general motor behaviours of the human body can be thought of as expressing emotional states. For instance,  the tilt angle of head can be linked inferiority- and superiority-related emotions~\cite{mignault2003many}.
Since such behaviour can be accounted for part of the cortical neural responses from neurotransmitter dopamine modulation (such as arousal), these bodily states tend to be relatively slow to arise and slow to dissipate. This facilitates people to differentiate emotions 
from observing these subtle behaviours. 

Conversely, similar as we do in our daily social communication, people are able to recognize others' emotions by their either static postures~\cite{walters1986perception} or dynamic motions \cite{walk1984emotion}. 
Studies suggested that there exist a hierarchical coding system~\cite{dael2012body,huis2014body,huis2014body2} 
or  `critical features'~\cite{roether2009critical} for emotion recognition.

From a technical perspective, to obtain a learning model about the relation between sensorimotor behaviours and emotions are also useful to endow an artificial agent (e.g. social robotic platform) 1) to obtain a natural non-verbal emotion expression, and 2) to recognize its counterpart's emotion.  
Thus, extended by the hierarchical sensorimotor architecture~\cite{zhong2015artificial}, we propose an emotion regulation model based on the perception-action model(PAM) theory~\cite{preston2002empathy}. 
A two-level neural learning model will be used to realize such an architecture. This neural model will show its feasibility in two different cases about motor action generation and personalized emotion recognition.

\section{A Bayeisan Emotion Regulated Perception-Action Model}
The framework of PAM is based on the common coding theory which advocates that action and perception are intertwined by sharing the same representational basis~\cite{prinz1997perception}. 
We asserted that this common representation between the perception and action is simply formed by either the mapping from  perception or the perceptual events that actions produce. 
Specifically, the representation does not explicitly represent actions; instead, there is an  encoding of the possible \emph{future} percept which is learnt from prior sensorimotor knowledge.
Therefore, 
this perception-action framework is derived from the ideomotor principle~\cite{james1890consciousness},
which advocates that actions are represented with prediction of their perceptual consequences, i.e. it encodes the forthcoming perception that is going to happen when an action is executed (i.e. motor
imagery) \cite{greenwald1970sensory}. 

 Due to the modulation role of emotion state, we propose that there is an intermediate level of internal state between the cognitive processes and the perception-action cycle. 
 As we stated, this level of internal state, such as emotion, regulates the expressions of sensorimotor behaviours.   
 To some extents, we can also regard this level of internal state as a level of common coding. As such, the representation of the internal state reflects the perception input and the motor imagination, forming an imagination of affective understanding such as empathy. 
Instead of the linkage between action and perception (e.g. sensorimotor contingency), the common coding theory proposes that perception and action may modulate each other directly via the shared coding by a similarity-based matching of common codes, even regulated by the internal state in our model. 
Therefore, the pairing of perception and action, i.e. the acquisition of  `common coding',  emerges from prior experience of the agent.
For instance, assume we have the hierarchical PAM model as a two-level architecture,  it is degenerated as the model stated in~\cite{preston2007perception}. 
If the agent (called `observer') observes that one person (called `presenter') is hurt by a knife,  the model may trigger involuntary and subtle movement when the agent is doing a certain kind of hand movement (e.g. moving a bit of the arm). Even the observer feels sympathy about the incident.
In this example, both of the current afferent information (referring to the perceived event) and predictive efferent information (referring to intended events from actions) have the same format and structure of the internal state representation.  

From the  aforementioned  sensorimotor functions, we propose a hierarchical cognitive architecture as shown in Fig.~\ref{fig: pam_mns} focusing on the emotion regulated functions to the sensorimotor behaviours. 
From the common coding framework of perception and action, the information of the feedback pathways are formed through various levels, regulated by internal states of the cognitive agent in the hierarchical architecture. 
Between the cognitive processes and the hierarchical motor controls, internal state (such as emotion) regulates perception-action cycle functions such as perception, motor imagery, and action planning.  
{ To establish the links between movements ($a$), the sensory effects ($e$) and the internal state ($s$), one may need one or more processes of latent learning~\cite{thistlethwaite1951critical}, reinforcement learning~\cite{wright1996reinforcement} or evolutionarily hard-wired acquisition~\cite{ledoux2007amygdala}. 
 This link, once establish, can be described in the following operations:

\begin{itemize}
   \item First, these associations allow to estimate the internal state, given the perceived behaviours performed by the others by means of the forward models (e.g. Bayesian Model) ($e \rightarrow s$). In the formulation of Bayesian inference, it can be written as Eq.~\ref{eq:bayeisan_sense}: 
   \begin{equation}
      P(S|A,E) \propto P(S)P(A|S)
  \label{eq:bayeisan_sense}
   \end{equation}
 where $S$ estimates the internal state evidence given an executed action $A$ (e.g. motor imagination by the observer itself), the perception $E$. 
 The term $P(A|S)$ suggests a pre-learnt internal model  representing the possibility of a motor action $A$ will be executed given a certain internal state ($S$). Since this model is solely by the observer itself, it may differ from various cultures,  ages, personalities. Therefore, learning of such an internal model is crucial. 
 Here another assumption is held that one's own action (or imagined action) $A$ is the same as its perceived action ($E$).

   \item Second, these associations allow the agent to move with a certain voluntary or involuntary behaviour given an internal state.  From the  backward computations introduced in Eq.~\ref{eq:bayeisan_action_cha4} ($s \rightarrow a$),  a predictive/generative sensorimotor integration occurs:
    \begin{equation}
      P(A|S) \propto P(A)P(S|E) + P(A)P(S|\cdot) 
  \label{eq:bayeisan_action_cha4}
   \end{equation}
   where $A$ indicates a motor action given the internal state $S$. In the equation we omit the factor of goal, but it is also the main target of any voluntary actions. Here we assume that one's internal state is determined by the current sensory input and a lot more factors ($P(S|\cdot)$).

  \item In terms of its hierarchical organization, it also allows this operation: with bidirectional information pathways, a low level perception representation can be expressed on a higher level, with a more complex receptive field, and vice versa ($e_{low} \leftrightarrow e_{high}$). 
  These operations can be achieved by extracting statistical regularity, which may be expressed as deep learning architectures.
\end{itemize}

\begin{figure}[h]
\centering
\includegraphics[width=1\linewidth]{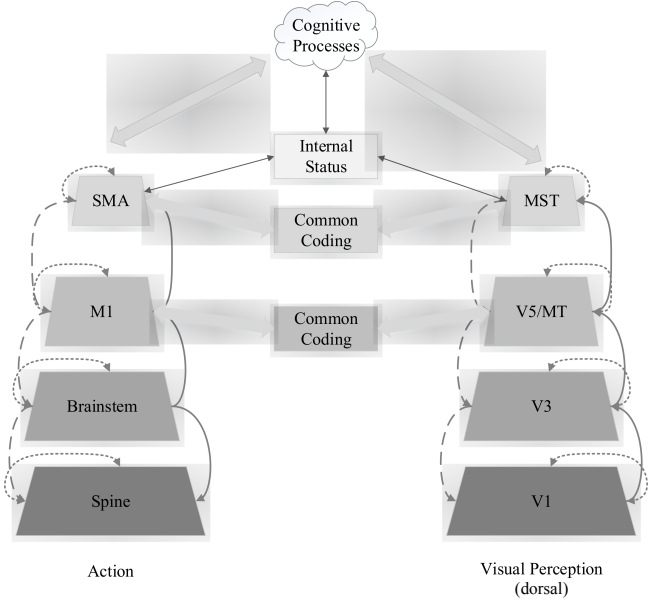}
\caption{A Hierarchical Perception-Action Model with Action and Visual Dorsal Stream}
\label{fig: pam_mns}
\end{figure}

Thus, this Bayesian framework proposes that  an internal state as one of the prior for feedback pathways 
represents a linkage between perception and motor actions. 
We will propose a simple learning model addressing the first two operations at the next section. After that we will demonstrate two case studies about its feasibility.

\begin{figure}
  \begin{center}
    \includegraphics[width=0.8\linewidth]{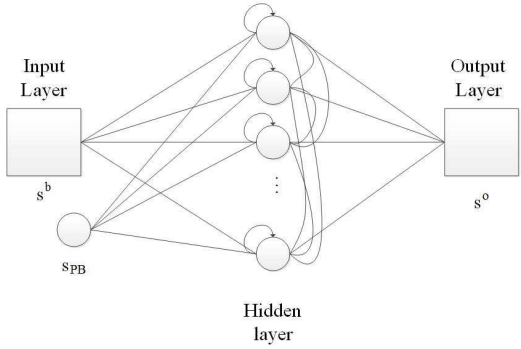}
    \caption{Recurrent Neural Network with Parametric Bias Units}
    \label{fig:srnpb}
  \end{center}
\end{figure}

\section{A Learning Model}
\subsection{Parametric Bias}
Recurrent Neural Network with Parametric Bias Units~\cite{tani2003self,tani2004self,zhong2011robot,zhong2014towards,zhong2014continuous} (Fig.~\ref{fig:srnpb}) is a family of recurrent neural networks that consist of an additional bias unit acting as bifurcation parameters for the nonlinear dynamics.  
In other words, the adjustable values in such a small number of units can determine different dynamics in the network with large dimension.  
Being different from ordinary biases, these units are trained from the training sequences with back-propagation through time (BPTT) in a self-organizing way, so the units exhibit the differences of the features of the training sequences. Furthermore, as  bifurcation parameters for nonlinear functions, the parametric bias units (PB units) endow the network to acquire the ability of learning different non-linear dynamic attractors, which makes it advantageous over generic recurrent networks.   

 Due to the above features, an RNNPB network can be regarded as a two-layer hierarchical structure for cognitive systems. Based on this feature, this network has also been further adopted to mimic the cognitive processes such as language acquisition~\cite{zhong2014towards}, language generalisation~\cite{sugita2005learning} and other cognitive processes.




\begin{figure*}[ht]
\centering
\subfloat[Learning Mode]{\includegraphics[width=0.31\textwidth]{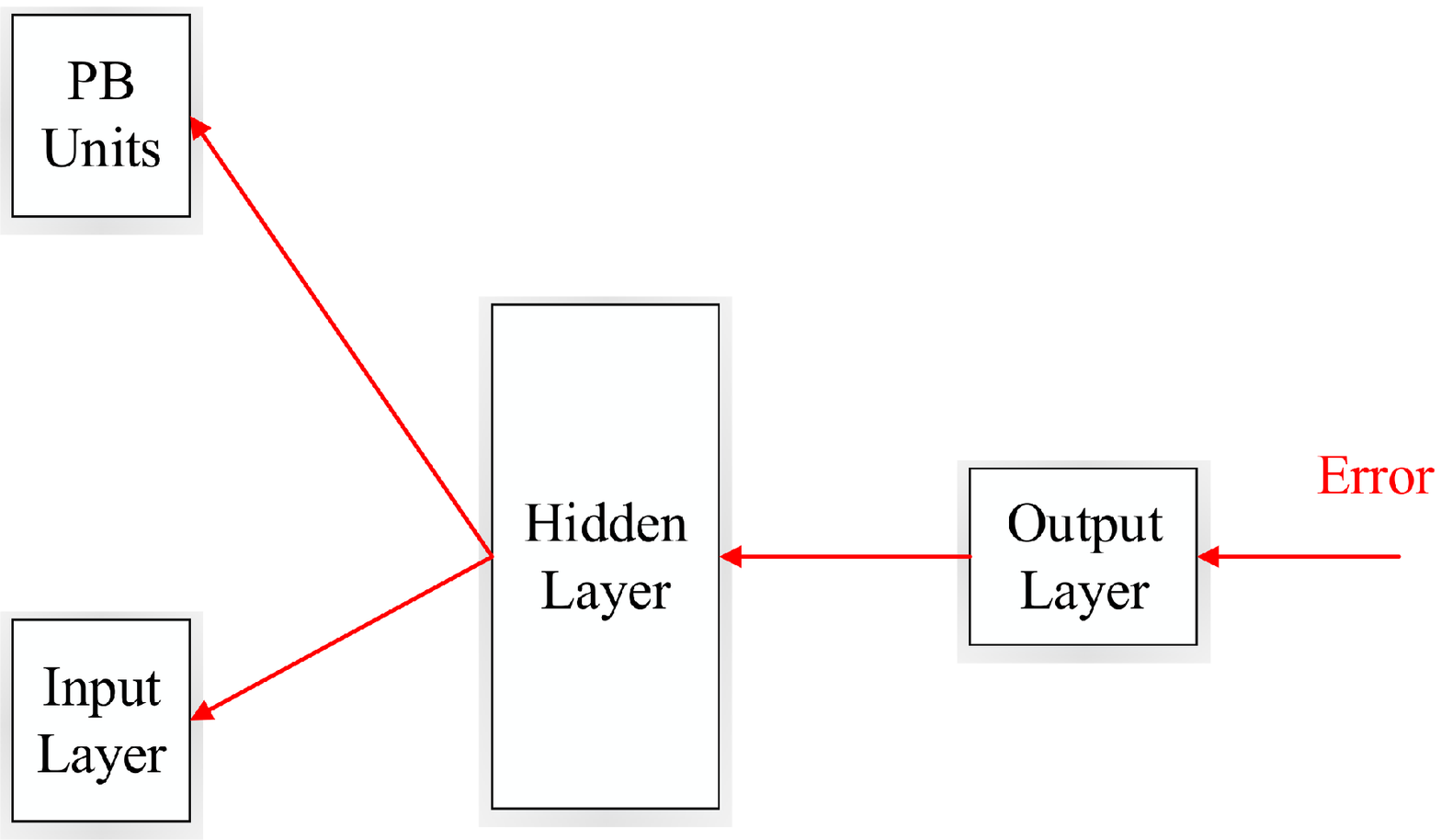}%
\label{fig:srn_learning}}
\hfil
\subfloat[Recognition Mode]{\includegraphics[width=0.31\textwidth]{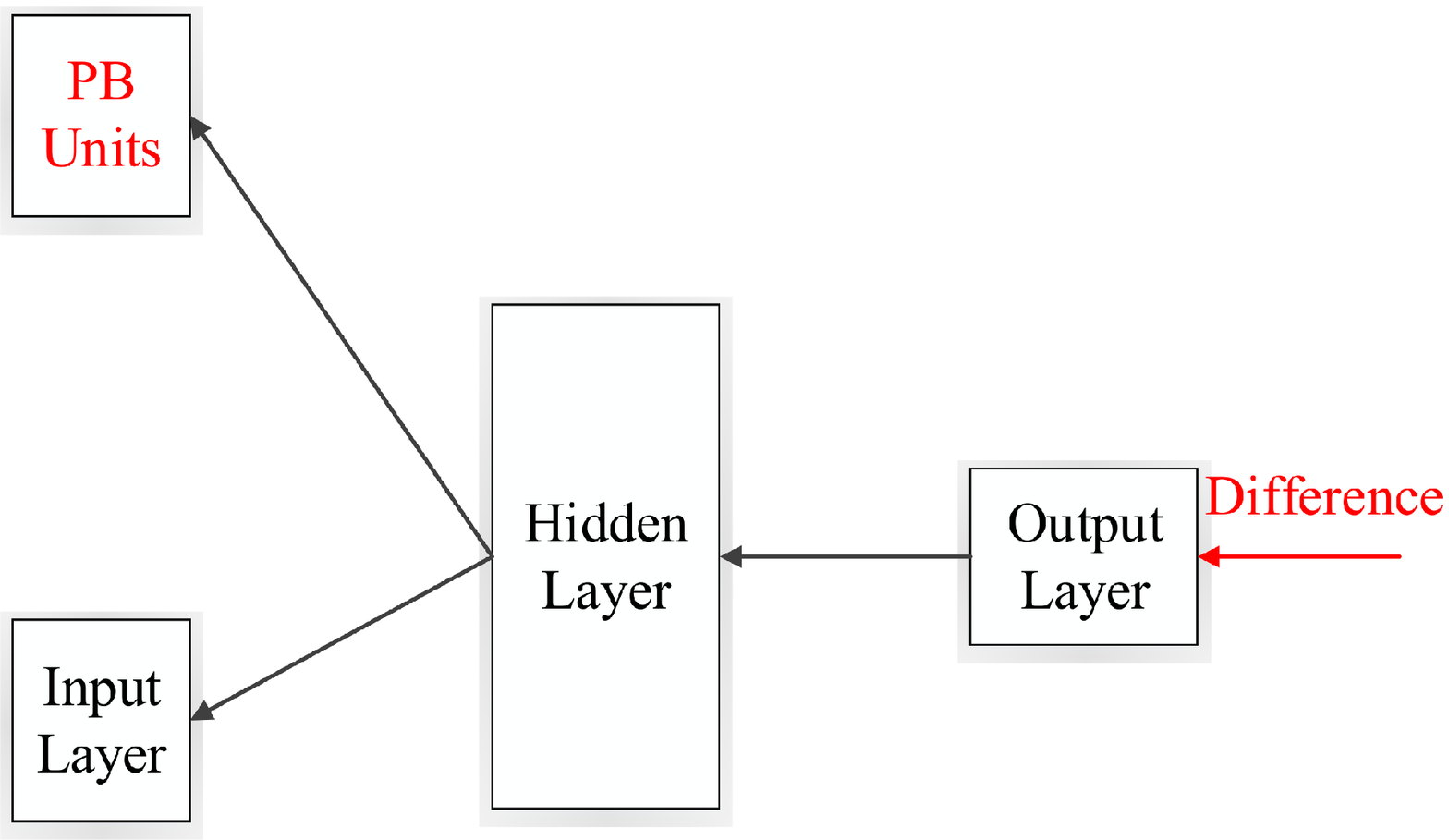}%
\label{fig:srn_recognition}}
\hfil
\subfloat[Generation Mode]{\includegraphics[width=0.31\textwidth]{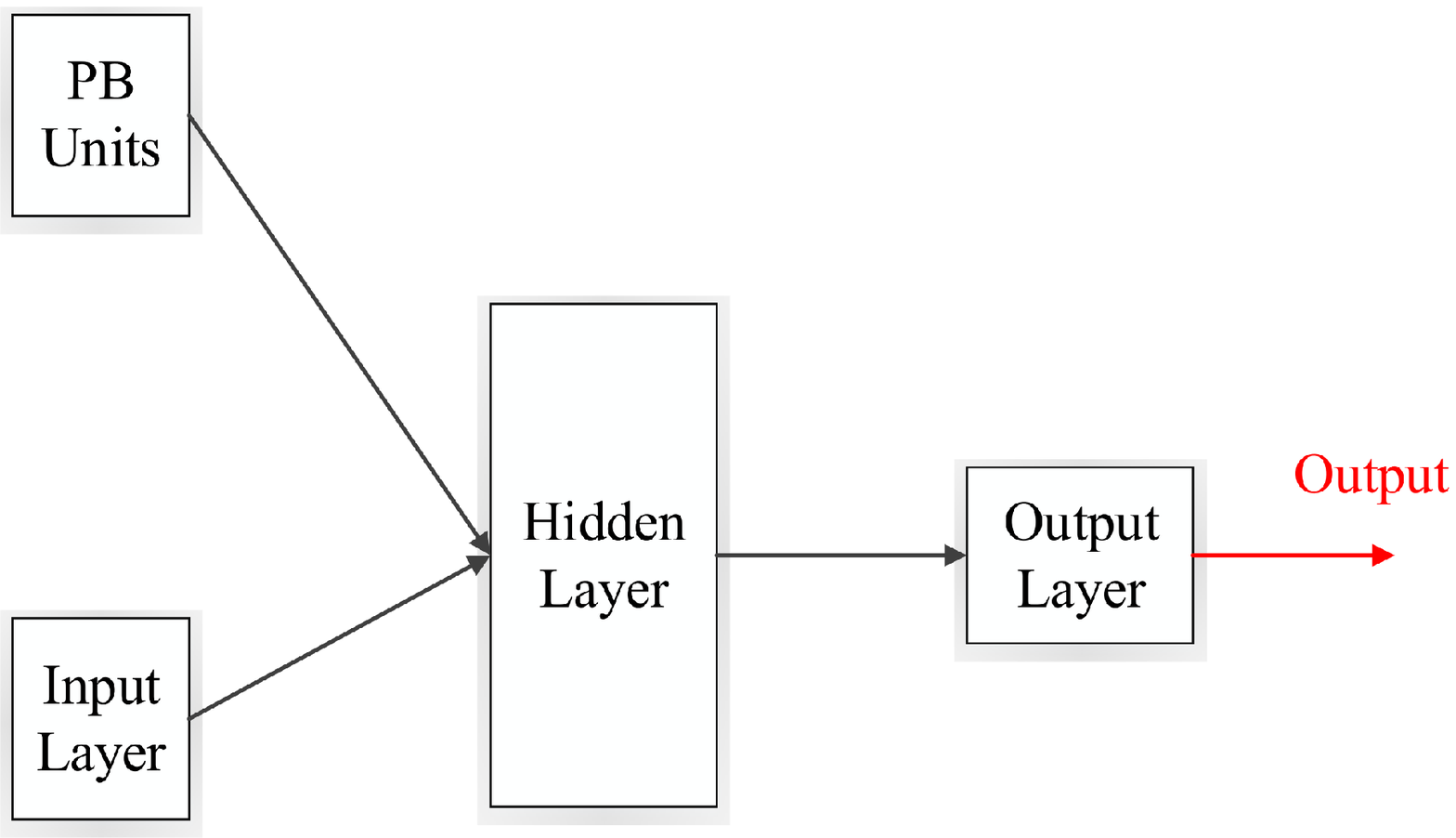}%
\label{fig:srn_prediction}}
\caption{Simulation results for the network.}
\label{fig_srn}
\end{figure*}

 \subsection{Three Modes}

 The RNNPB  three running modes in an RNNPB network, namely learning, recognition and generation modes. They functionally simulate different stages between sensorimotor sequences and high-level internal states of these sequences. The neural dynamics and algorithms of these modes are as following:
 
 \begin{itemize}
   \item Learning mode  (Fig.~\ref{fig:srn_learning}): The learning algorithm performs in an off-line and supervised way as generic recurrent neural networks do. When providing the training stimulus with a new pattern, the weights connecting neurons are updated with BPTT (back-propagation through time). Besides, the residual error from the BPTT updates the internal values in PB units; the internal values of the PB units are updated in a self-organising way by this error derived from the back-propagation. To achieve the slowness update of the internal value,  in each epoch $e$, the $k$th PB unit $u$ updates its internal value based on the summation of the back-propagated error from one complete sequence.
   
   \item Recognition mode (Fig.~\ref{fig:srn_recognition}):
In this mode, the network recognises which type of sequences by updating the internal values of PB units. The information flow in this mode is mostly the same as in the learning mode, i.e. the error is back-propagated from output neurons to the hidden neurons, but the synaptic weights are not updated; rather, the back-propagated error only contributes to the updating of the PB units. By this mean, if a trained sequence is presented to the network, the values of the PB units will converge to the ones that were previously shown in the learning mode in order to restore the PB values trained before. 

\item Prediction mode (Fig.~\ref{fig:srn_prediction}): After learning and after the synaptic weights are determined, the RNNPB can act in a closed-loop way: the output prediction can be used as an input for the next time step. In principle, the network can generate a trained sequence by providing initial value of the input and externally setting the PB values.  
   
 \end{itemize}

\subsubsection{Learning Mode}
During learning mode, if the training progress is basically determined by this cost function $C$; following gradient descent, each weight update in the network is proportional to the negative gradient of the cost with respect to the specific weight $w$ that will be updated:
\begin{equation}
\Delta w_{ij} = -\eta_{ij} \frac {\partial C}{\partial w_{ij}}
\end{equation}

\noindent where $\eta_{ij}$ is the adaptive learning rate of the weights between neuron $i$ and $j$, which is adjusted in every epoch. To determine whether the learning rate has to be increased or decreased, we compute the changes of the weight $w_{i,j}$ in consecutive epochs:
\begin{equation}
\sigma_{i,j} = \frac{\partial C}{\partial w_{i,j}}(e-1)
\frac{\partial C}{\partial w_{i,j}}(e)
\end{equation}

The update of the learning rate is 
\begin{eqnarray}
\eta_{i,j}(e) = \left\{ \begin{array}{rl}
min(\eta_{i,j}(e-1) \cdot \xi^+, \eta_{max}) & \mbox {if $ \sigma_{i,j} > 0$}, \\
max(\eta_{i,j}(e-1) \cdot \xi^-, \eta_{min}) & \mbox {if $ \sigma_{i,j} < 0$}, \\
\eta_{i,j}(e-1) & \mbox{else.} \end{array} \right. 
\end{eqnarray}
\noindent where $\xi^+ > 1 $ and $\xi ^- < 1$ represent the increasing/decreasing rate of the adaptive learning rates, with $\eta_{min}$ and $\eta_{max}$ as lower and upper bounds, respectively. Thus, the learning rate of a particular weight increases by $\xi^+$ to speed up the learning when the changes of that weight from two consecutive epochs have the same sign, and vice versa. 

As mentioned before, besides the usual weight update according to back-propagation through time, the accumulated error over the whole time-series also contributes to the update of the PB units. The update for the $i$-th unit in the PB vector for a time-series of length $T$ is defined as:

\begin{equation}
\label{eq: PB-error}
\rho_i (e+1) = \rho_i (e) + \gamma_i \sum_{t=1}^T \delta_{i,j}^{PB}\\
\end{equation}

where $\delta^{PB}$ is the error back-propagated to the PB units, $e$ is $e$-th time-step in the whole time-series (e.g. epoch), $\gamma_i$ is PB units' adaptive updating rate which is proportional to the absolute mean value of the back-propagation error at the $i$-th PB node over the complete time-series of length $T$:

\begin{equation}
\gamma_i \propto \frac{1}{T} {\sum_{t=1}^T \delta_{i,j}^{PB}}
\end{equation}

The reason for applying the adaptive technique is that it is hard for the PB units to converge in a stable way. Usually a smaller learning rate is used in the generic version of RNNPB to ensure the convergence of the network. This results in a trade-off in convergence speed. However, the adaptive learning rate we used is an efficient technique to overcome this trade-off.

\subsubsection{Recognition Mode}
The recognition mode is executed with a similar information flow as the learning mode: given a set of spatio-temporal sequences, the error between the target and the real output is back-propagated through the network to the PB units. However, the synaptic weights remain constant and only the PB units will be updated, so that the PB units are self-organized as the pre-trained values after certain epochs. Assuming the length of the observed sequence is $a$, the update rule is defined as:   

\begin{equation}
\label{eq:PB_recognition}
\rho_i (e+1) = \rho_i (e) + \gamma \sum_{t=T-a}^T \delta_{i,j}^{PB}
\end{equation}

\noindent where $\delta^{PB}$ is the error back-propagated from a certain sensory information sequence to the PB units and $\gamma$ is the updating rate of PB units in recognition mode, which should be larger than the adaptive rate $\gamma_i$ at the learning mode.

\subsubsection{Generation Mode}
The values of the PB units can also be manually set or obtained from recognition, so that the network can generate the upcoming sequence with one-step prediction.

Furthermore, according to~\cite{ito2004generalization}, a trained RNNPB not only  can  retrieve and recognise different types of pre-learnt, non-linear oscillation dynamics, as an expansion of the storage capability of working-memory within the sensory system, but also adds the generalisation ability to recognise and generate untrained non-linear dynamics and compositions.

 \section{Case Studies}
 \subsection{Case 1: Generation of Non-verbal Emotion Expression by Avatar Simulation}

 In this demonstration, we used a simulation based on RNNPB with the parameters shown in Tab.~\ref{tab:parameter} to generate  specialised non-verbal robot behaviours.  
   We separated two kinds of training data based on two different behaviours in order to eliminate the subtle differences in different behaviours generation, corresponding we omit the goal $G$ in Eq. \ref{eq:bayeisan_action_cha4}.
  
The training data was taken from an inertial motion capture system by Xsens\footnote{www.xsens.com/en/general/mvn}, while the actor was performing two sets of basic behaviours (standing and walking) expressing five kinds of emotions (joy, sadness, fear, anger, pride).
This three-dimensional skeleton motion capture was used instead camera-based systems such as Microsoft Kinect, because it would be easier to map the generated motion into humanoid robots for further project developments. Also motions from various body parts could be well-isolated in a quantitative way for network training. Such quantitative data can be reviewed and utilised in external editors and programs (Fig. \ref{fig:sample-body-tracker}). The sampling frequency of this system was 120Hz.

\begin{figure}[ht]
  \centering
    \includegraphics[width=0.3\textwidth]{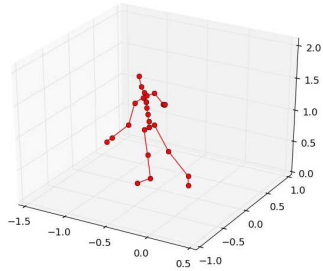}
    \caption{Three Dimensional Skeleton}
    \label{fig:sample-body-tracker}
 \end{figure}

After training, as depicted in Eq.~\ref{eq:bayeisan_action_cha4}, given that the PB values we obtained from training, we can generate the pre-trained or un-trained temporal sequences in the network's generation mode. 
These behaviours were represented in a 78-dimensional data-set by reconstructing the network output. 
These simulations were done by adjusting the PB units into the pre-trained values and the \emph{novel} values as the midpoints between either two pre-trained values in the PB space.
In the avatar demonstrations, only the body skeleton were shown by means of the lines connecting to skeleton points. 
Part of the demonstrations can be viewed online\footnote{http://youtu.be/9yahOKcEi-A}. 
Apart from the trained emotions, some `interval' emotions from the behaviours can be also perceived when the novel points in the PB spaces were selected.

\begin{table*}[th]
	\centering
	\begin{tabular}[c]{c|c|c}
		\hline
		Parameters & Parameter's Descriptions & Value\\
		\hline
		$\eta$ & Initial Learning Rate& $2.0 \times 10^{-6}$ \\ 
		$ \eta_{max} $ & Maximum Value of Case 1 & $1.0 \times 10^{-4}$ \\
		$ \eta_{min} $ & Minimum Value of Case 1 & $1.0 \times 10^{-8}$ \\
		$ \eta_{r} $ & Updating Rate in Case 2 & $8.0 \times 10^{-3}$ \\
		$ M_{\gamma} $ & Proportionality Constant of PB Units Updating Rate & $0.001$ \\
			$n_h$ & Size of Hidden Layer & $100$ \\
		$n_{PB} $ & Size of PB Unit  & $2$ \\
		$\xi^-$ & Decreasing Rate of Learning Rate & $0.999999$ \\
		$\xi^+$ & Increasing Rate of Learning Rate & $1.000001$ \\
		\hline
	\end{tabular}
	\caption{Network parameters}
	\label{tab:parameter}
\end{table*}

\subsection{Case 2: Personalised Emotion Recognition}
In the second demonstration, the recognition mode of the network was investigated  (Eq.~\ref{eq:bayeisan_sense}) using the parameters shown in Tab.~\ref{tab:parameter}.
Here we will also testify the \textit{real-time}  recognition ability of this method. 
Training was done by capturing the Kinect tracking data by OpenNI from one person. 
We only selected the upper body of the data (9-dimension: head, neck, torso, left shoulder, left elbow, left hand, right shoulder, right elbow, right hand),  since the emotion is more significant in the upper body. 
After training, the Kinect captured the data again to test whether the networks can distinguish the emotion correctly. 
The real-time demonstration video\footnote{http://youtu.be/JusCuKvHg44} showed that the PB value has a bit delay depending on the choice the updating rate. 
However, such a delay can be used as a confidence level to simulate the decision processes for the robot.   

To quantitatively evaluate the performance of the learning, we also compare  the differences between the original PB value and the recognised PB value. 
During recognition, a stopping criterion is set if the updating of PB values was smaller than a threshold $0.1$ in consecutive $100$ times. Tab.~\ref{tab:dist-standing} listed the distance of PB values under the training and recognition modes. 
Although during the experiments we could not guarantee the training and recognition data-sets were identical, 
we can still observe that behaviours from the same emotion drove the PB values closer to each other. Furthermore, we can also notice the behaviours from `joy' and `pride', `fear' and `anger' were also quite close which may indicate some connections between these two pairs of emotions. 

\begin{table*} 
 \centering
   \begin{tabular}{c|ccccc}
    \hline
    \backslashbox{Train}{Rec.} & joy & pride  & fear & anger & sadness   \\
    \hline
      joy & 0.3587 & 0.9755 & 3.5893 & 3.2201 &  5.8812\\
       
      pride & 0.9074 & 1.0143 & 3.5741 &  4.1577 & 7.6670  \\
       
      fear & 3.1214 & 4.0362 & 0.9783 & 1.2778 & 3.0431 \\
       
      anger & 4.2789 & 3.6609 & 1.2128 & 0.9955 & 9.1021  \\
       
      sadness & 4.6976 & 7.2737 & 3.4684 & 9.4351 & 0.8954  \\
       
      \hline 
    \end{tabular} \caption{Differences between PB Values(Standing Behaviour)}
 \label{tab:dist-standing}
\end{table*}

\section{Summary}
In this paper, we proposed a cognitive framework focusing on the regulation role of emotions to sensorimotor behaviours. 
This hierarchical architecture is based on the perception-action model, in which the internal state (emotion) acts as a prior in the feedback pathways to further modulate perception and actions. 
Following the proposed two Bayesian inferences between internal state and the sensorimotor behaviours, 
we propose to use a recurrent neural network with parametric bias units (RNNPB) to contruct a two-layer architecture in which the parametric bias represents an internal state. 
The non-verbal expression generation and emotion recognition demonstrations witness the feasibility of this hierarchical architecture. In the next steps, a few possibilities can be made to further extend this architecture: 1) more factors, such as goal, perception from previous states and prior knowledge, can be considered in the Bayesian inferences; 2) a more sophisticated hierarchical learning method can be used to extract and generate more statistical regularities; 3) we will further extend this hierarchical PAM model with ASMO architecture\cite{novianto2014flexible} to allow a flexible attention between various conditions as well as modulated by internal status such as emotion.

\bibliography{Bibliography2}
\bibliographystyle{abbrv}

\end{document}